\documentclass{ieeeaccess}
\usepackage{cite}
\usepackage{amsmath,amssymb,amsfonts}
\usepackage{algorithmic}
\usepackage{graphicx}
\usepackage{textcomp}

\usepackage{cite}
\usepackage{amsmath,amssymb,amsfonts,bm}
\usepackage{algorithm}
\usepackage{algorithmic}
 
\usepackage{graphicx}
\usepackage{textcomp}
\usepackage{subcaption}
\usepackage{comment}

\def\BibTeX{{\rm B\kern-.05em{\sc i\kern-.025em b}\kern-.08em
    T\kern-.1667em\lower.7ex\hbox{E}\kern-.125emX}}
\begin{document}

\doi{https://doi.org/10.1109/ACCESS.2022.3163270}

\title{Revisiting Bayesian autoencoders with MCMC}
\author{\uppercase{Rohitash Chandra}\authorrefmark{1,3}, \IEEEmembership{Senior Member, IEEE},
\uppercase{Mahir Jain  \authorrefmark{2},   Manavendra Maharana ,
Jr}.\authorrefmark{1,3}, Pavel N. Krivitsky \authorrefmark{1}}
\address[1]{ Transitional Artificial Intelligence Research Group, School of Mathematics and Statistics, UNSW Sydney, NSW 2052, Australia}
\address[2]{Manipal Institute of Technology, Karnataka, India}
\address[3]{UNSW Data Science Hub, UNSW Sydney, NSW 2052, Australia}

\markboth
{Author \headeretal: Preparation of Papers for IEEE TRANSACTIONS and JOURNALS}
{Author \headeretal: Preparation of Papers for IEEE TRANSACTIONS and JOURNALS}

\corresp{Rohitash Chandra: First A. Author (rohitash.chandra@sydney.edu.au).}

\begin{abstract} 

Autoencoders gained popularity in the deep learning revolution given their ability to compress data and provide dimensionality reduction. Although prominent deep learning methods have been used to enhance autoencoders, the need to provide robust uncertainty quantification remains a challenge. This has been addressed with variational autoencoders so far. Bayesian inference via Markov Chain Monte Carlo (MCMC) sampling has faced several limitations for large models; however,  recent advances in parallel computing and advanced proposal schemes have opened routes less traveled. This paper presents Bayesian autoencoders powered by MCMC sampling implemented using parallel computing and Langevin-gradient proposal distribution. The results indicate that the proposed Bayesian autoencoder provides similar performance accuracy when compared to related methods in the literature. Furthermore, it provides uncertainty quantification in the reduced data representation. This motivates further applications of the Bayesian autoencoder framework for other deep learning models.

\end{abstract}


\titlepgskip=-15pt

\maketitle

\section{Introduction} \label{sec:introduction}

Autoencoders are a family of unsupervised learning methods that use neural network architectures and learning algorithms to learn a lower-dimensional (reduced) representation   which can then be used to reconstruct a representation close to the original data. They thus facilitate dimensionality reduction for prediction and classification problems \cite{baldi2012autoencoders,doersch2016tutorial}. Autoencoders  have been successfully applied to image classification \cite{luo2018convolutional,zhou2019learning}, face recognition \cite{
gao2015single,kan2014stacked}, geoscience and remote sensing \cite{dong2018review}, speech-based emotion recognition \cite{sahu2018adversarial}, and data generation \cite{ranjan2018generating}. Autoencoders have been prominent in deep neural network architectures and also for transfer learning tasks \cite{masci2011stacked,deng2013sparse}. Recent developments  include regularized autoencoders \cite{alain2014regularized}, variational autoencoders \cite{kingma2013auto,
doersch2016tutorial, sonderby2016ladder, tolstikhin2017wasserstein,liang2018variational}, adversarial autoencoders \cite{zhao2018adversarially,beggel2019robust,sahu2018adversarial,pidhorskyi2018generative,creswell2018denoising}, variational graph autoencoders \cite{kipf2016variational,ding2020variational}, and convolutional autoencoders \cite{masci2011stacked}.

Bayesian neural networks incorporate alternative training via Bayesian inference in which the model parameters (weights and biases) are (jointly) represented by a probability distribution rather than point estimates given by conventional gradient-based training methods \cite{mackay1995probable}. The \emph{posterior} probability distribution naturally accounts for uncertainty in parameter estimates, which is further propagated into the decision-making process \cite{neal2012bayesian,neal2011mcmc}. Bayes' theorem is used as the foundation for inference in Bayesian neural networks. Markov Chain Monte Carlo (MCMC) sampling methods \cite{mackay1992practical} implement Bayesian inference by estimating the posterior distribution, which is often referred to as sampling from the posterior distribution. Variational inference \cite{hinton1993keeping} provides another way to approximate the posterior distribution, which approximates an intractable posterior distribution by a tractable one. This makes it particularly suited to large data sets and models, and so it has been popular for autoencoders and neural networks \cite{kingma2013auto,graves2011practical}. Variational autoencoders \cite{kingma2013auto} employ variational inference \cite{hinton1993keeping} as a form of regularisation of the model parameters (weights) during  training   \cite{doersch2016tutorial, sonderby2016ladder, tolstikhin2017wasserstein,liang2018variational}.  

 A deep learning model can feature tens of thousands to millions of parameters that are typically optimized by gradient-based algorithms.  The major challenge of MCMC methods for training (sampling) deep learning models is the computational time required for constructing a good approximation of the posterior distribution, incurring heavy computational costs. However,  progress in incorporating gradient-based proposals into MCMC sampling has been inspired by earlier works  \cite{
roberts1996exponential,roberts1998optimal}  for the development of Hamiltonian and Langevin-based MCMC methods \cite{neal2011mcmc,neal2012bayesian,welling2011bayesian}. Langevin-gradient proposal distribution demonstrated to be effective for Bayesian neural learning \cite{Chandra2019NC, Chandra2020NC} and combining it with parallel computing via tempered MCMC resulted in further improvements \cite{Chandra2019NC}, which motivates their application for deep autoencoders that can feature thousands of model parameters.

This paper presents Bayesian autoencoders powered by tempered MCMC sampling that incorporate parallel computing and  Langevin-gradient proposal distribution. We demonstrate the effectiveness of Bayesian autoencoders with benchmark datasets that involve small-scale to large-scale models featuring a maximum of about a million parameters. We also investigate the effect of two prominent gradient-based training methods and develop an adaptive Langevin-gradient proposal distribution for tempered MCMC.  We finally compare our results with the literature where gradient-based methods have been used for the same datasets.  The  Bayesian autoencoder would provide a  principled approach to uncertainty quantification in reduced data representation using MCMC. The Bayesian autoencoder can be used to reconstruct an ensemble of the reduced dataset from the posterior distribution rather than a single one given by single-point estimates given by gradient-descent learning in conventional autoencoders. Hence, once the posterior distribution is obtained, different estimators for the Bayesian autoencoder’s parameters are possible such as the  \textit{maximum a posteriori probability} (MAP) estimate \cite{bassett2019maximum}. 

The rest of the paper is organized as follows. In Section~\ref{sec:review}, we present a background and literature review of related methods. Section~\ref{sec:methodology} presents the proposed methodology, followed by experiments and results in Section~\ref{sec:experiments}. Section~\ref{sec:discussion} provides a discussion, and Section~\ref{sec:conclusion} concludes the paper with directions of future work.

\section{Related work} \label{sec:review}

\subsection{Autoencoders}

Autoencoders compress data by providing a reduced feature set that may be used for supervised learning. Autoencoders have been used with supervised learning methods  such as simple   neural networks \cite{betechuoh2006autoencoder}, recurrent neural networks \cite{d2017autoencoder},   long-short term memory (LSTM) networks \cite{zhao2017robust}, and  graph neural networks \cite{taheri2018learning}. Depending on the dataset, the features from other deep learning methods are used to augment the canonical autoencoder. For instance, a deep convolutional autoencoder incorporates pooling and convolutional layers used in convolutional neural networks for unsupervised learning tasks \cite{autounsupervised}. 
The applications of autoencoders in machine learning problems include data classification, dimensionality reduction, pattern recognition, image denoising, anomaly detection, and recommender systems \cite{appae, ZHAO202010, JI2020, ZIHAO2020, ZHANG2020, LI2020}. Autoencoders have been successfully applied to a number of real-world applications such as cyber-security \cite{yousefi2017autoencoder}, medical imaging \cite{gondara2016medical}, and biometrics \cite{d2017autoencoder}. Furthermore, a review of autoencoders has been done for geoscience and remote sensing with a focus on hyperspectral images  \cite{dong2018review}.

Next, we review some of the key autoencoder architectures from the literature \cite{varia}. Sparsity autoencoders \cite{sparse} are used to capture the latent, rather than redundant, representations of the data using a loss function that penalizes activation of the hidden layer; hence, enforcing the sparsity constraint. Denoising autoencoders \cite{denoising} involve adding artificial noise to the data, after which the autoencoder attempts to reconstruct the original data from noisy data. This prevents the autoencoder from capturing an overly-complete representation of the data which is common to many conventional autoencoders. Contractive autoencoder \cite{contractive} makes the reduced representation less sensitive to small variations in the input samples; this is achieved by adding a regularizer to the loss function which reduces the reduced representation's sensitivity to the input. Convolutional autoencoders \cite{conae, WANG202010}  create a reduced representation of image data which is helpful for supervised machine learning.

\subsection{Bayesian deep learning }

The major limitation of conventional autoencoders is the lack of uncertainty quantification in the estimation of model parameters and the challenge of  tuning  hyperparameters. Bayesian neural networks address these by estimating the parameters of the model (weights and biases) as random variables, in contrast to single point estimates by backpropagation learning that employs gradient-based methods \cite{mackay1992practical,mackay1995probable}. The progress in neural networks has led to a revolution in deep learning models; however, progress in Bayesian neural networks and Bayesian deep learning has been relatively slow due to certain challenges \cite{wang2020survey,polson2017deep}. Bayesian inference implemented with canonical MCMC methods used to  face major difficulties in handling the large number of model parameters. Variational inference provides an alternative Bayesian inference approach with successful  implementations in  variational autoencoders \cite{kusner2017grammar}, generative adversarial networks (GANs) \cite{mescheder2017adversarial}, variational convolutional neural networks (CNNs) \cite{gal2015bayesian,zhao2019variational,chien2017variational,zhou2020variational}, variational recurrent neural networks via LSTM networks \cite{su2018variational}, and variational graph neural networks \cite{bonner2019temporal,kipf2016variational,qu2019gmnn}. Further details about their applications are given in \cite{wang2020survey}. We note that variational autoencoders also use a probabilistic representation of model parameters, as opposed to a single value representation in conventional autoencoders, facilitating uncertainty quantification \cite{vae, XU202010, ZHU202012}.  Finally, the importance weighted autoencoder extends the variational autoencoder to use multiple samples to approximate more complex posteriors \cite{burda2015importance}.

 The limitations of MCMC sampling have been addressed with better computational resources and advanced proposal distributions, incorporating gradients \cite{welling2011bayesian,neal2012bayesian,neal2011mcmc}. Hamiltonian MCMC sampling methods have been used for Bayesian neural networks \cite{levy2017generalizing} with enhanced computation strategies
 \cite{cobb2020scaling}. Langevin-based MCMC methods have been successfully used in the implementation of Bayesian neural networks for pattern classification and time series prediction problems \cite{chandra2019langevin}.  A major challenge in MCMC sampling has been in addressing big data and computationally expensive models; hence, surrogate-assisted estimation, in which a low-cost surrogate model provides an approximation of the likelihood, has been used to address this issue \cite{chandra2020surrogate,llorente2021survey}. Langevin-based MCMC methods have also been used in Bayesian neural networks for transfer learning given multiple sources of data \cite{chandra2020bayesian}.   In previous research, simple Bayesian neural networks have been used that at most had several thousand model parameters  \cite{chandra2020surrogate,chandra2020bayesian}; hence,   the challenge lies in adapting them for autoencoder-based deep learning models that can feature up to a million parameters.

\section{Methodology} \label{sec:methodology}


\subsection{Model and Priors}

An autoencoder consists of two major parts: an encoder function $f_\phi(\mathbf{x})$ and a decoder function $ \bar{f}_\psi(\mathbf{h})$, where $\mathbf{x}$ is the data that represents a set of features, and $\mathbf{h}$ is a set of  reduced (latent) features.  The autoencoder is assessed by how well the decoder can reconstruct the data from the encoder using a loss function $R_{\text{loss}}$ as given 
\begin{equation}
 R_{\text{loss}}= \operatorname*{arg\,min}_{\phi,\psi}\, \lvert\mathbf{x}-(\bar{f}_\psi (f_\phi(\mathbf{x})) \rvert^2 
\end{equation}

 where $\phi$ and $\psi$ represent the parameters (weights and biases) of the encoder and decoder, respectively.

In particular, $z(\mathbf{x},\theta)$ represents a feedforward neural network, with $\theta$ being its set of weights and biases. An encoder-decoder pair is then constructed from it, as given below

\begin{align} 
\mathbf{h}&= f_\phi(\mathbf{x}) = z(\mathbf{x},\phi)\\ \nonumber
 \mathbf{x} &= \bar{f}_\psi(\mathbf{h}) = z(\mathbf{h},\psi).
\end{align}

\begin{figure*}[htpb!]
 \centering
 \includegraphics[width=13cm]{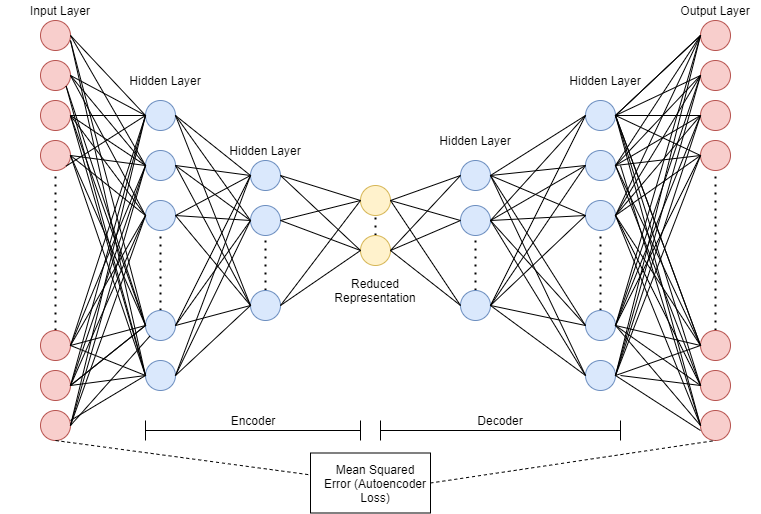}
 \caption{\label{fig:auto-encoder}The autoencoder features the encoder, the decoder which are highlighted.}
\label{fig:ae}
 \end{figure*}

The encoder extracts the essential features into a reduced representation while the decoder is used to reconstruct the input from the reduced representation as shown in Figure~\ref{fig:auto-encoder}. The challenge is in determining the optimal $\theta=(\phi,\psi)$ and typically gradient-based learning algorithms are used to minimize $R_{\text{loss}}$ \cite{baldi2012autoencoders,doersch2016tutorial}. In case the data are labeled, another way to measure the performance of the autoencoder is by training another neural network model using the features from the reduced representation ($\mathbf{h}$), and comparing it with performance when trained with original data. In contrast to principal component analysis (PCA), autoencoders are capable of nonlinear data reduction and can detect repetitive structures in the data \cite{wang2016auto}.

Our likelihood function compares the original data $\mathbf{x}$ with decoder output $\mathbf{x}' =z(\mathbf{x}, \theta) = z(z(\mathbf{x}, \phi), \psi)$. We incorporate the $\tau^2$ parameter in the model which denotes the variance of data, $\tau^2$ is  also estimated; hence, our combined set of parameters becomes $\Theta= (\theta, \tau^2)$.  We only consider continuous data for the Bayesian autoencoder, where mean-squared-error (MSE) is used as the loss function   \cite{Autoencoder-MSELoss}. Hence, we  assume the data to have Gaussian distribution and use a Gaussian likelihood  given by
\begin{equation}
P(\mathbf{x}|\Theta)=\frac{1}{(2\pi\tau^2)^{N/2}}
 \exp\left(-\frac{1}{2\tau^2}\sum_{t=1}^N\left(y_t-\bar{E}(x_t|{\bf x}'_t)\right)^2\right),
\label{lhoodreg}
\end{equation}

where $\bar{E}(x_t|{\bf x'}_t)$ is the output of the neural network model for $x_t$  input features  denoted by $t$ instances in the training data.   $N$ is the total number of instances in the training data. The priors are based on Gaussian distribution for $\theta$ and inverse gamma distribution for $\tau^2$, respectively. Hence, the autoencoder weights and biases ($\theta$) are independent \emph{a priori} with a normal distribution with zero mean and variance $\sigma^2$. If $\theta$ features $L$ model parameters in total, its joint prior with $\tau^2$ is
 \begin{multline}
\bar{P}(\Theta) \propto \frac{1}{(2\pi\sigma^2)^{L/2}}
 \exp\Bigg\{-\frac{1}{2\sigma^2}\bigg(\sum_{l=1}^L \theta_l^2 \bigg) \Bigg\}\\
\times \tau^{2(1+\nu_1)}\exp\left(\frac{-\nu_2}{\tau^2}\right),
\label{prior_regression}
\end{multline}
 where, $\nu_1$ and $\nu_2$ are user chosen constants. We update $\tau^2$   via random-walk proposal distribution and use the log-scale to avoid numerical  instabilities in computing the prior and likelihood.

\subsection{Langevin-gradient Metropolis--Hastings}

 We now present the MCMC sampler for training the model parameters (posterior) of the Bayesian autoencoder. We use a combination of methods: 1.) efficient proposal distribution that uses Langevin-gradients, 2.) parallel computing, 3.) efficient multimodal sampling via tempered MCMC (parallel tempering).  We utilize the Langevin-gradient proposal   distribution that incorporates Gaussian noise with gradients for a single iteration (sample).   We note that Langevin-gradient proposal distribution has  been effective  for novel Bayesian neural networks in relatively small and large neural network architectures of up to several thousand model parameters
\cite{Chandra2019NC, ChandraK2020NC,Chandra2020EngAI,chandra2021bayesiangraph}; however, in this work we will experiment with deep autoencoders that feature up to a million model parameters.

The Langevin-gradient (LG)  proposal distribution  for a given sample   ($n$) proposes to update the vector of weights and biases $\theta_n$ from a biased multivariate normal distribution 
\begin{equation}
 \theta_{n}^{\star}\sim\mathcal{N}(\theta_n+\nu_3 \times\nabla E( \theta, \bf x  ), \nu_4^2 I_L) \label{update}
\end{equation}
where $\nu_3$ (learning rate) and $\nu_4^2$  are user-defined tuning parameters.
$I_L$ is an $L \times L$ identity matrix used to generate the stochastic noise. We take  into account the loss function  $E( \theta, \bf x)$, for the autoencoder model $f()$ that features data $\bf x$ as shown below
\begin{equation}
E(\theta, \bf x)  =   \sum_{t\in{\bf x}}(\bf x_t -f({\bf x}_t, \theta) )^2.
 \end{equation}
Hence, the  gradient   $\nabla E( \theta, \bf x)$ is given by
 
\begin{equation}
\nabla E( \theta, \bf x) =
\left(\frac{\partial{E}}{\partial{\theta_1}},\ldots, 
\frac{\partial{E}}{\partial{\theta_{L}}}\right).
\end{equation}
 
 The proposal attempts to ``explore'' the posterior density, given   our prior $\bar{P}(\Theta)$ and likelihood  $P(\mathbf{x}|\Theta)$ defined in Equations \eqref{prior_regression} and   \eqref{lhoodreg}. Furthermore, with $\Theta^\star_n=(\theta^\star_n,\tau^2)$, we either accept or reject the proposal using the standard Metropolis-Hastings criterion
\begin{equation}
\alpha = \min\bigg\{1, \frac{P(\textbf{x}|\Theta_{n}^{\star})\bar{P}(\Theta_{n}^{\star})Q(\Theta_{n}|\Theta_{n}^{\star})}{P(\textbf{x}|\Theta_{n})\bar{P}(\Theta_{n})Q(\Theta_{n}^{\star}|\Theta_{n})} \bigg\},\label{eq:MH}
\end{equation}
with $Q(\Theta_{n}^\star|\Theta_{n})=P(\Theta_{n}^\star|\Theta_{n})$, the conditional proposal density and vice versa. In general, $\nabla_\theta \log \{\bar{P}(\Theta_n)P(\mathbf{x}|\Theta_n)\}\ne\nabla_\theta \log \{\bar{P}(\Theta_n^\star)P(\mathbf{x}|\Theta_n^\star)\}$ and so $Q(\Theta_{n}^\star|\Theta_{n})\ne\ Q(\Theta_{n}|\Theta_{n}^\star)$ -- an asymmetric proposal, and thus they do not cancel in Equation \eqref{eq:MH}.

\subsection{Adaptive Langevin-gradient proposal distribution}

In practice, in our trial experiments, we found that using the  LG proposal distribution directly produces relatively slow mixing leading to inferior results for the case or large neural networks such as autoencoders. Hence, we formulate a new proposal distribution, borrowing ideas from the popular Adam optimizer \cite{kingma2014adam}.
The Adam-based weight update in the conventional learning paradigm is expressed as follows

\begin{align}
\theta_{k} &= \theta_{k-1}- \alpha   \times \hat{g}   \\ 
\hat{g} &= \frac{\sqrt{1-\beta_2^k}}{\sqrt{1- \beta_1^k}} \cdot \frac{\eta_{k}}{\sqrt{\mu_{k}}+\epsilon} \nonumber
\label{gradientadam}
\end{align}

where
$$\eta_{k}=\beta_1 \eta_{k-1}+(1-\beta_1)   \times \nabla E( \theta, \bf x)_{k}$$
$$\mu_{k}=\beta_2 \mu_{k-1}+(1-\beta_2)  \times  \nabla E( \theta, \bf {x})_{k}^2$$


where, \textcolor{black}{$\beta_1$ and $\beta_2$ are first and second moment estimates, respectively}.  Moreover,  $\epsilon$ is a small scalar used to prevent division by 0. $\alpha$ is the user defined learning rate and typically, the default values are  $\beta_1=0.99$, $\beta_2=0.999$, $\alpha=10^{-3}$, and $\epsilon=10^{-8}$ (implementation in PyTorch \footnote{https://pytorch.org/docs/stable/optim.html}).

In our previous work \cite{chandra2021bayesiangraph}, Adam-based gradients have been utilized in adaptive Langevin gradient (adapt-LG) proposal distributions for convolutional graph neural networks, which produced better results when compared to LG proposal distribution. We note that the learning rate is a user-chosen hyperparameter that is fixed during sampling in the case of LG and adapt-LG proposal distributions. Stochastic gradient descent (SGD) does not adapt the gradients while Adam uses a heuristic to adapt the gradients during optimization. Hence, Adam-based updates make use of earlier values of the gradient, which makes the Markov transition kernel in MCMC time-inhomogeneous.
In our Bayesian autoencoder framework, we present an enhanced version known as optimization adapt-LG* which only makes use of adapt-LG in the burn-in phase and switches to LG afterwards. In this way, we address the inhomogeneous nature of adapt-LG.  Hence, we use an enhanced form of gradients $\hat{E}( \theta, \bf x  )$  to obtain the adapt-LG proposal distribution as follows

 \begin{equation}
 \hat{\theta}_{n}\sim\mathcal{N}(\theta_n+\nu_3 \times \hat{g}, \nu_4^2 I_L) \label{adamprop}
\end{equation}

where $\nu_3$ is user-defined  learning rate previously denoted as $\alpha$.

\subsection{Tempered MCMC Framework}

\begin{figure*}[tb]
\centering
 \includegraphics[width=190mm] {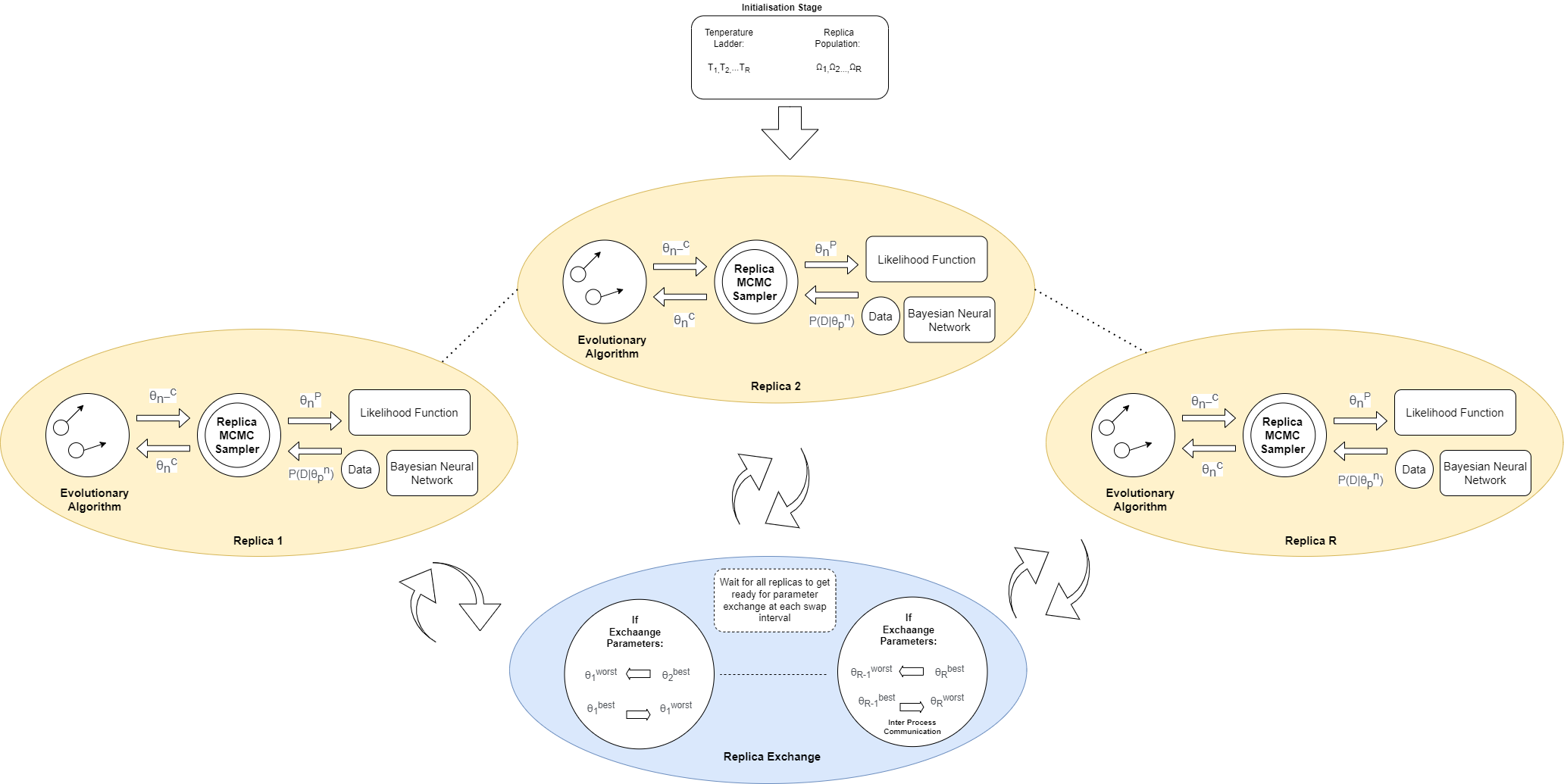}

\caption{Bayesian autoencoder framework highlighting tempered MCMC utilizing parallel computing and autoencoder neural network. 
}

\label{fig:framework}
\end{figure*}

The development of \emph{tempered MCMC} (also called \emph{parallel tempering} and \emph{replica exchange MCMC}) has been   motivated by the cooling behaviour of certain materials in the field of thermodynamics \cite{swendsen1986replica,hukushima1996exchange,hansmann1997parallel}. In tempered MCMC, an ensemble of $M$ replicas $\Omega = [R_1, R_2, \dotsc, R_M]$, is executed with a distinct temperature value $T= [1,\dotsc, T_\text{max}]$, where $T_\text{max}$  defines the extent of exploration. A replica defined by temperature $t$, samples from an attenuated posterior distribution $P_t(\Theta_{n}^t|\textbf{x})\propto P(\textbf{x}|\Theta_{n}^t)^{1/t}\bar{P}(\Theta_{n}^t)$. This serves to ``flatten'' the distribution, giving replicates with high $t$ a generally higher probability of accepting proposals. The ensemble of replicas  facilitates enhanced exploration capabilities, in particular, it enables a given replica to escape from a local maximum, allowing multi-modal and discontinuous posteriors to be explored \cite{sen1996bayesian,maraschini2010monte}.

Inferentially, this can be viewed as augmenting the sampler state $\Theta_n$ into $\Theta_n^t=(\theta,\tau^2,t)$ and sampling from them jointly with those realisations, given $t=1$ features the target distribution. Therefore, periodically, a proposal can exchange with the states of the neighbouring replicas, so that $\Theta_{n}^{t\star}=\Theta_{n}^{t'}$ and $\Theta_{n}^{t'\star}=\Theta_{n}^{t}$, with $t'=t+1$, which is accepted with the probability
\begin{equation}
\beta = \min\bigg\{1, \frac{P(\textbf{x}|\Theta_{n}^{t})^{1/t'}P(\textbf{x}|\Theta_{n}^{t'})^{1/t}}{P(\textbf{x}|\Theta_{n}^{t'})^{1/t'}P(\textbf{x}|\Theta_{n}^{t})^{1/t}} \bigg\},\label{eq:PT}
\end{equation}
since the proposal is symmetric and the priors cancel.

 We note that in the temperature ladder $T$, not only the $T_{\text{max}}$ is important, but also the distribution of the temperature values. Specifically, for the geometric path between prior and posterior, ideally one should take temperatures such that the sum of the Kullback-Leibler (KL) divergences between successive tempered distributions is minimized.  However, we use a geometric temperature ladder that worked well for Bayesian neural networks from our previous research \cite{ChandraK2020NC,chandra2021bayesiangraph}.

\subsection{Bayesian Autoencoder}

We now present the details for implementing a tempered MCMC framework for Bayesian autoencoders that utilize parallel computing.  Figure~\ref{fig:framework} presents the framework that features the autoencoder model, tempered  MCMC replica ensemble, and the inter-process communication for the exchange of neighboring replica states.  

  We present further details in Algorithm 1, where we first define the autoencoder topology by specifying the number of inputs, hidden layers, and outputs for the encoder,  and  the decoder (in reversed order). Secondly, we define the hyperparameters for the tempered MCMC sampler, such as the number of  replicas ($M$),  geometric temperature ladder -  maximum temperature ($T_\text{max}$), replica swap interval ($R_\text{swap}$) that determines how often to propose an exchange between neighboring replicas, and the maximum number of samples (iterations) for each replica ($R_\text{max}$). In parallel processing, it is difficult to determine when to stop sampling, and hence this is done by checking several alive (active) replica processes. Therefore, we need to set the number of replicas in the ensemble as $alive$; $alive = M$. These are the key steps in initialization as shown in Stage 0 of Algorithm 1. The algorithm uses tempered MCMC in its first phase, then switches to canonical MCMC in its second phase, which is defined by several samples that needs to switch $R_\text{switch}$. During the switch, all the temperature values in the ladder are changed to 1 as done in our earlier works \cite{Chandra2019NC,chandra2019PT-Bayeslands}.   The algorithm implements adapt-LG* proposal distribution (LG in the first phase, and adapt-LG in the second phase) using $R_\text{switch}$.   

We then begin within replica sampling (Stage 1.1) by iterating through the respective replicas, where the manager process executes and manages parallel processes. In Stage 1.2, each replica sample creates a proposal using either random-walk or Langevin-gradient  (either LG or adapt-LG*) (Equation \eqref{update}) and computes the likelihood (Stage 1.3) and uses Metropolis-Hastings condition in Stage 1.4 to check if the proposal is good enough to be part of the posterior distribution. In Stage 1.5, the algorithm checks if it needs to use tempered MCMC or needs to transform into canonical MCMC where temperature values of all the replicas in the ensemble are set to 1. Hence,  tempered MCMC employs adapt-LG and canonical MCMC employs LG proposal distribution. Note that the samples obtained by tempered MCMC are later discarded as part of the burn-in period. In Stage 2.1, the algorithm computes the likelihood of the replica-exchange which is similar to the way it's computed in Equation \eqref{eq:PT}, and then it exchanges the neighboring replica using the Metropolis-Hastings condition (Stage 2.2). Note that we represent the likelihood  in log-scale to avoid numerical instabilities. 

In Stage 3, the algorithm checks if   ($R_\text{max}$) has been reached, and if so, the number of replicas alive is decremented to eventually stop the entire sampling process. Finally, we reach the end in Stage 4 where the sampling process ends, and we combine the respective replica posterior distributions of the Bayesian autoencoder for further analysis.

 Rather than swapping the entire replica state (weights and biases in the autoencoder), it is possible to swap the respective neighboring replica temperature values, which is cheaper computationally.  This is done by simply swapping neighing replicate temperature values, which is useful in parallel computing implementation, where inter-process communication can be memory intensive and computationally costly.  

\begin{figure}[htpb!]
 \centering
  \includegraphics[width=9cm]{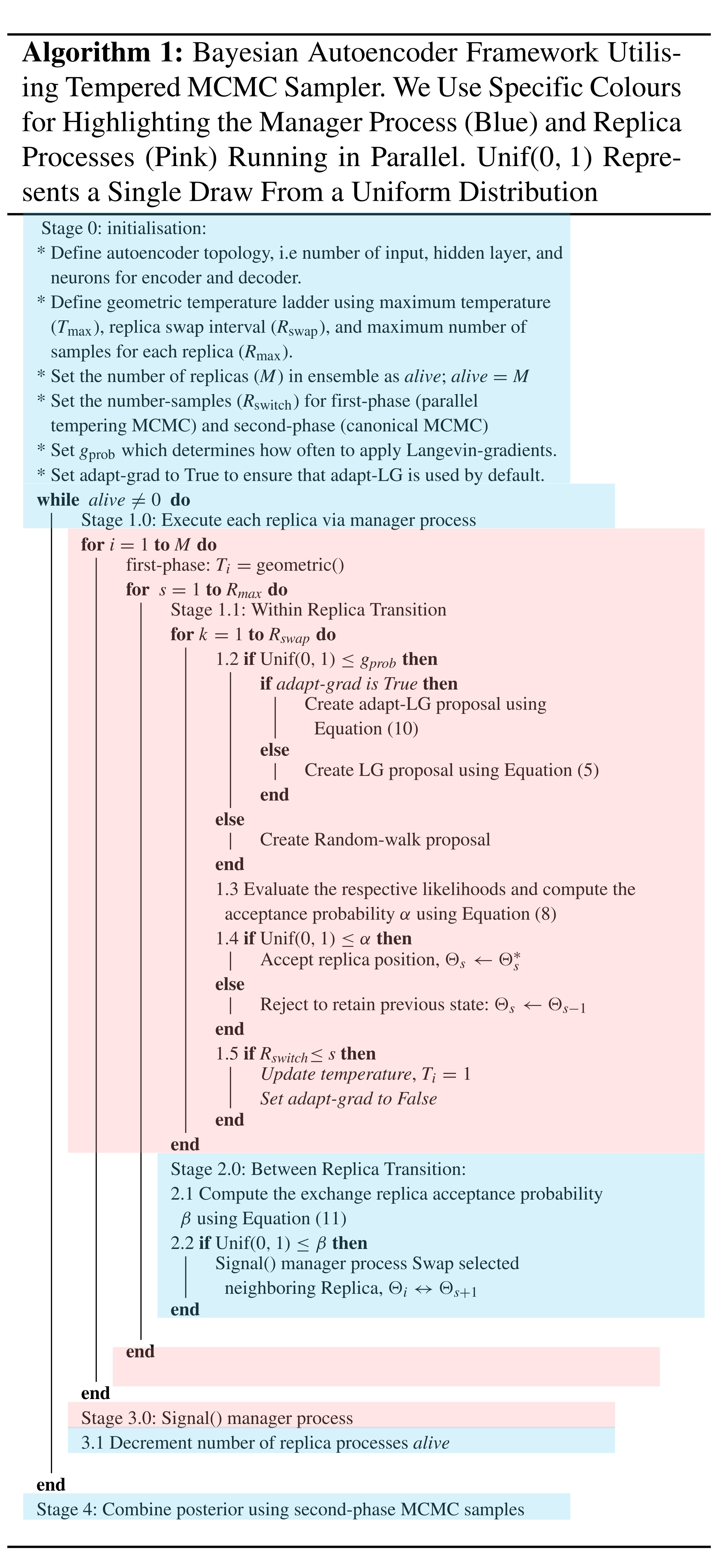}

\label{alg:ptfnn}
 \end{figure}

\section{Experiments and Results} \label{sec:experiments}

\subsection{Dataset Description}


We select benchmark datasets commonly used in the literature of autoencoders and data reduction methods. The datasets feature distinct properties in terms of size, features, and application. These include Madelon \cite{Madelon}, Coil-2000 \cite{Coil2000} and Swiss Roll \cite{SwissRoll} which are summarised in Table~\ref{dataset}.

  Madelon is a synthetic dataset that features data points grouped in 32 clusters, each on a vertex of a five-dimensional hypercube. The clusters are randomly labeled +1 or -1. In addition to these 5 coordinates, 15 linear combinations have been added for a total of 20 (redundant) informative features, as well as 480 random distractors. Coil-2000 has 86 variables containing customer information, including product usage data and socio-demographic data derived from postal codes, and whether the customer has a caravan insurance policy. Lastly, the synthetic Swiss Roll dataset features data points in the shape of a Swiss Roll given in three dimensions.  It is a popular dataset for visualisation of data reduction methods \cite{SR_1,SR_2,SR_3}.

\begin{table}
\caption{\label{tab:Table 1} Overview of dataset used for dimensional reduction.}
\label{dataset}
\centering
 \begin{tabular}{c c c c}
 \hline
 Dataset & No. Features & No. Train & No. Test \\
 \hline
 Madelon & 500 & 2000 & 1800 \\
 Coil-2000 & 85 & 5822 & 4000 \\
 Swiss Roll & 3 & 3750 & 1250 \\

 \hline
\end{tabular}
\end{table}
 
 The respective datasets were normalized before model training in the range of [0,1]. This was done to ensure that all features had equal importance in the model.

\subsection{Implementation}

 Table~\ref{tab:topology} provides the Bayesian autoencoder (adapt-LG*) topology that is used in the experiments described in this section. In the respective datasets, the input feature size and the feature size of the compressed data are different. Table~\ref{tab:topology} highlights the exact dimensions of the data at the input, intermediate, and output layers. Table~\ref{tab:topology} also demonstrates the sheer volume in the number of parameters being dealt with and its exponential rise related to the increase in the number of input features  from the data.

We evaluate some of the critical parameters in   MCMC sampling to observe the effects on the sampling performance, computational time, and reconstruction accuracy. The Bayesian autoencoder is designed to reduce the Madelon dataset from 500 features to 300. In the case of the Coil 2000 dataset, the Bayesian autoencoder reduces 85 features to 50 features. In the Swiss Roll dataset, the Bayesian autoencoder reduces 3 features to 2 features, essentially ``unwrapping'' the Swiss Roll, as shown in Figure~\ref{fig:SRVisualization}. 

\begin{table*}[htbp!]
 \caption{Bayesian autoencoder (adapt-LG*) topology showing the total number of parameters (weights and biases).}
\label{tab:topology}
\centering
\small
\begin{tabular}{c c c c c c c c }
\hline
Data & Input Layer & Hidden Layer 1 & Hidden Layer 2 & Hidden Layer 3 & Hidden Layer 4 & Output Layer & Total Parameters\\
& (Input--Output) & (Input--Output)&(Input--Output) &(Input--Output) &(Input--Output) &(Input--Output)
\\
\hline
Madelon & 500--450 & 450--400 & 400--300 & 300--400 & 400--450 & 450--500 & 1053701\\
Coil 2000 & 85--70 & 70--60 & 60--50 & 50--60 & 60--70 & 70--85 & 27037\\ 
Swiss Roll& 3--10 & 10--5 & 5--2 & 2--5 & 5--10 & 10--3 & 224 \\

\hline
\end{tabular}

\end{table*}

In all the experiments, we use the tempered MCMC sampler hyper-parameters as follows. In random-walk proposals and the random component of the Langevin proposal, the Gaussian noise is added with a standard deviation $\nu_4 = 0.005$, centered at 0. We use $\nu_1 = 0$ and $\nu_2 = 3$ as shown in the prior distribution given in Equation \eqref{prior_regression}. We use replica temperature ladder maximum $T_\text{max}=2$, and swap interval $R_\text{swap}=5$ samples  for all experiments. We use $M=8$ replicas with $R_\text{max}=6000$ MCMC samples per replica. After $R_\text{switch}=3000$ samples, tempered  MCMC converts to canonical MCMC with neighborhood exchange by setting the replica's temperature to 1. Note that we chose  $R_\text{max}$ in trial experiments by observing the trend and selecting a value, where the accuracy and likelihood do not have major changes. In terms of $R_\text{swap}$, we selected a value from trial experiments that ensured that a good mixing of the replicas take place. 

  Finally, we compare our results with several common classification algorithms: $K$-nearest neighbors (KNN) \cite{knn}; support vector machines (SVM) \cite{SVM}; random subspaces ensemble (RSE) \cite{EnsembleClass} method which uses ensemble of decision trees; and naive Bayes \cite{singh2020multiobjective}. We implement the Bayesian autoencoder using the pyTorch library \footnote{pyTorch library: \url{https://pytorch.org/}} with Python multi-processing library \footnote{Python multi-processing library: \url{https://docs.python.org/3/library/multiprocessing.html}} for parallel tempered MCMC replica processes.

\subsection{Preliminary investigation: parameter tuning}

\begin{figure*}[htb]
 \centering
 \begin{subfigure}{0.3\textwidth}
 \centering
 \includegraphics[scale = 0.35]{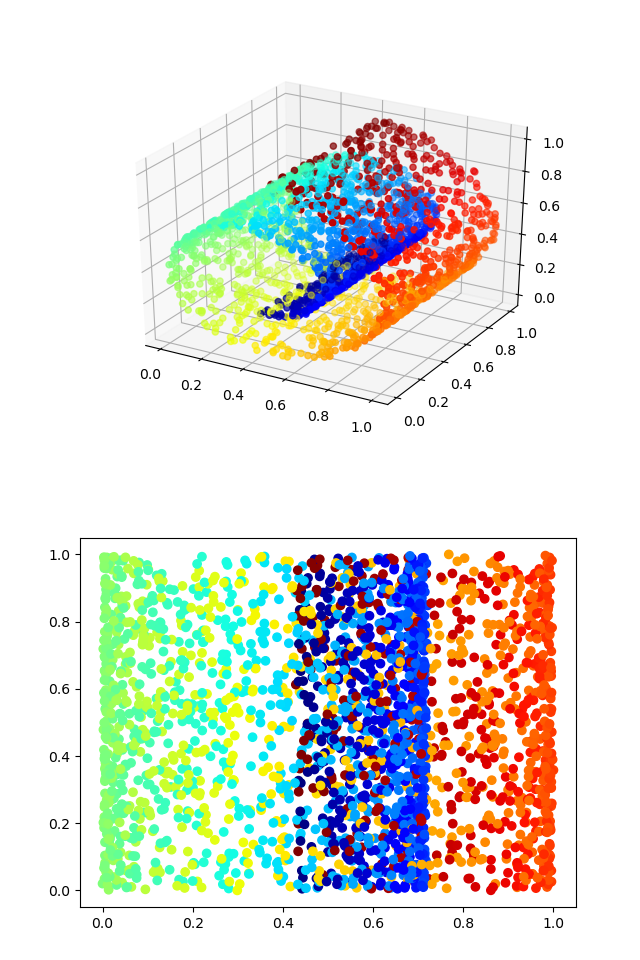}
 \caption{Original\label{subfig:SRV-original}}
 \end{subfigure}
 \begin{subfigure}{0.3\textwidth}
 \centering
 \includegraphics[scale = 0.35]{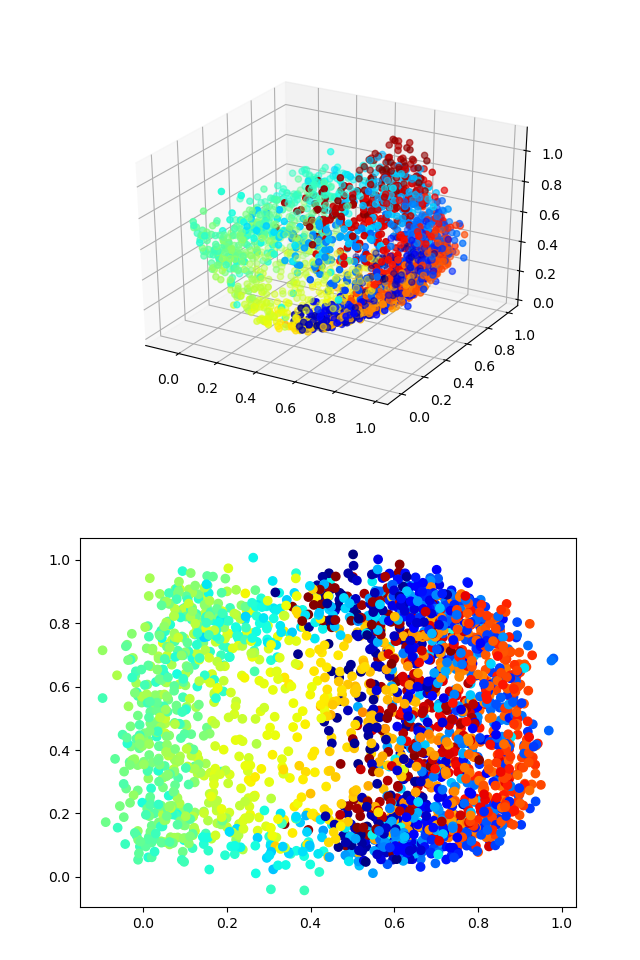}
 \caption{Canonical Autoencoder (MSE=0.012)\label{subfig:SRV-canon}}
 \end{subfigure}
 \begin{subfigure}{0.3\textwidth}
 \centering
 \includegraphics[scale = 0.25]{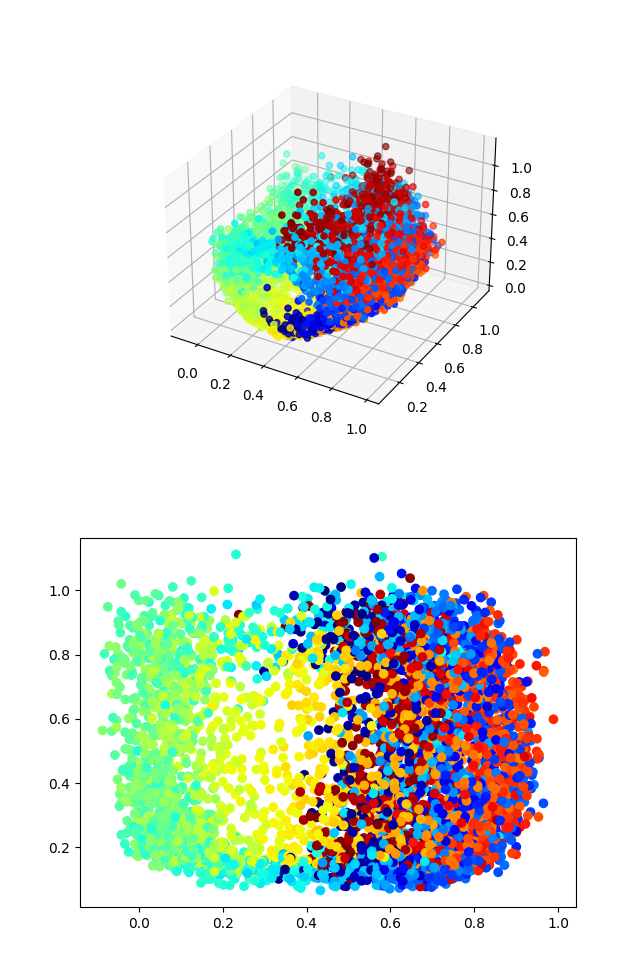}
 \caption{Bayesian Autoencoder (MSE=0.010)\label{subfig:SRV-bayesian}}
 \end{subfigure}
 \caption{Swiss Roll Visualization for different autoencoders }
 \label{fig:SRVisualization}
\end{figure*}

We first carry out a preliminary investigation to understand the effect of certain parameters on the tempered MCMC sampling and show a visualization of the results using the Swiss Roll dataset.
 Table~\ref{tab:SR_MSE} gives the accuracy for different Bayesian autoencoder configurations in MCMC sampling with adapt-LG* proposal distribution
 for the Swiss Roll dataset. We show the trend of the reconstruction accuracy (MSE) and acceptance percentage  as we change the MCMC hyper-parameters, i.e. step-size (variance) in proposal distribution, and learning rate for Langevin-gradients.  The highlighted configuration in Table~\ref{tab:SR_MSE} is visualised in Figure~\ref{fig:SRVisualization}. We also find that the accuracy for the chosen configuration is close to the canonical autoencoder. However, we note that the step size of 0.03 with learning rate of 0.04 provides the best performance, since they provide a balance between accuracy (0.025) and acceptance performance (19.25 \%). In MCMC sampling, only the  accuracy is not desired, a good acceptance rate is also needed to ensure that the sampler has well explored the posterior distribution.  We note that in some models an acceptance of around 23.40 \% \cite {BEDARD20082198} is known to be optimum for MCMC sampling.

 Figure~\ref{fig:SRVisualization} gives a visual representation of the Swiss Roll dataset for 2500 data points. It shows the 3D and 2D views of the various stages of the visualization process. Panel~(\subref{subfig:SRV-original}) shows the original dataset; Panel~(\subref{subfig:SRV-canon}) shows the reconstructed dataset after reducing and then reconstructing it through a canonical autoencoder; and Panel~(\subref{subfig:SRV-bayesian}) uses the same architecture but the Bayesian autoencoder. The visualization and analysis (Table~\ref{tab:SR_MSE}, and Figure~\ref{fig:SRVisualization}) demonstrates the effectiveness of the Bayesian autoencoder and its applicability to larger datasets.

\begin{table*}[htbp!]
 \caption{Reconstruction accuracy (MSE) with different settings of tempered MCMC parameters for the two instances of the Swiss Roll dataset using Topology shown in Table~\ref{tab:topology}.}
\label{tab:SR_MSE}
\centering
\small

\begin{tabular}{c c c c c c}
\hline
Method & Topology & MSE & Acceptance Percentage & Step-size & Learning-rate \\
\hline
Canonical Autoencoder (Adam) & Swiss Roll
& 0.012
& -
& -
& -
\\
Bayesian Autoencoder (adapt-LG*) & Swiss Roll
& 0.025
& 19.250
& 0.030
& 0.040
\\
\textbf{Bayesian Autoencoder (adapt-LG*)} & Swiss Roll
& \textbf{0.010}
& \textbf{0.125}
& \textbf{0.005}
& \textbf{0.010}
\\
Bayesian Autoencoder (adapt-LG*) & Swiss Roll
& 0.104
& 31.630
& 0.080
& 0.090
\\

\hline
\end{tabular}

\end{table*}

\begin{table*}[htbp!]
\caption{Reconstruction accuracy (MSE) as a function of the adapt-LG* rate for the Madelon dataset. } 
\label{tab:result_lgrate}
\centering
\small
\begin{tabular}{c c c c c c c}
\hline
adapt-LG* rate &Train MSE &Test MSE &Swap \% &Acceptance \% &Time (mins.)\\
 & Best (Mean,Std) & Best (Mean,Std) & & &\\
\hline
 0
&  0.946 (1.908,0.952)
& 0.937 (1.899,0.951)
& 78.14
& 1.20
& 184

 \\
 0.25
& 0.019 (0.024,0.004)
& 0.019 (0.025,0.004)
& 77.23
& 15.78
& 242

 \\
 0.5
& 0.011 (0.027,0.004)
& 0.013 (0.027,0.004)
& 78.39
& 22.68
& 302

 \\
 0.75
& 0.012 (0.021, 0.007)
& 0.013 (0.023,0.006)
& 71.57
& 34.64
& 359
 \\
 1
& 0.012 (0.019, 0.005)
& 0.012 (0.021,0.002)
& 76.85
& 40.25
& 435

 \\

\hline
\end{tabular}
\end{table*}

\begin{table*}[htbp!]
 \caption{Effect of the number of replicas on Madelon dataset using adapt-LG* proposal distribution.}
 
\label{tab:replicas}
\centering
\small
\begin{tabular}{c c c c c c c}
\hline
 Replicas &Train MSE &Test MSE &Swap \% &Acceptance \% &Time (mins.)\\
 & Best (Mean,Std) & Best (Mean,Std) & & & \\
\hline
 2
& 0.012 (0.018, 0.009)
& 0.013 (0.019, 0.006)
& 27.29
& 32.0
& 564

 \\
 4
& 0.013 (0.020, 0.01)
& 0.015 (0.023, 0.008)
& 42.67
& 32.5
& 527

 \\
 6
& 0.013 (0.014,0.008)
& 0.014 (0.016, 0.007)
& 56.93
& 32.0
& 483

 \\
 8
& 0.012 (0.021, 0.007)
& 0.013 (0.023,0.006)
& 71.57
& 34.64
& 359

 \\

\hline
\end{tabular}

\end{table*}

\subsection{Results}

We evaluate specific aspects of our proposed Bayesian autoencoder framework using the Madelon dataset. The Bayesian autoencoder reduces the number of the features from 500 to 300 features (encoder) and then reconstructs it, i.e.  from the 300 features into the 500 features (decoder). We report the accuracy between the reconstructed dataset and the original dataset. We first present results in terms of computational efficiency, the effect of adapt-LG* rate, and several replicas ($M$)  for the Madelon dataset. We note that the Langevin-gradient is for one iteration or epoch which is defined by a single forward and backward pass via backpropagation.

 Table~\ref{tab:result_lgrate} shows the effect of adapt-LG* rate ($g_{\text{prob}}$) on the performance of the Bayesian autoencoder given by training and testing accuracy (MSE) in terms of best, mean, and standard deviation (std). Since all of these are sampling from the same posterior, the accuracy should be the same as well. However,  we find that without the adapt-LG* (when the rate is 0), the sampler cannot find the posterior mode. On the other hand, adapt-LG* proposals are much more computationally costly as given by the time in minutes (mins.).

 Next, we evaluate the effect of the number of replicas ($M$) using the Bayesian autoencoder using adapt-LG* proposal distribution. Table~\ref{tab:replicas} summarises the results where we observe a lower computational time with a similar accuracy (MSE) as the number of replicas increases, suggesting that parallel processing is useful in this case.

\begin{figure*}[htbp!]
 \centering
 \begin{subfigure}{0.3\textwidth}
 \centering
 \includegraphics[scale = 0.33]{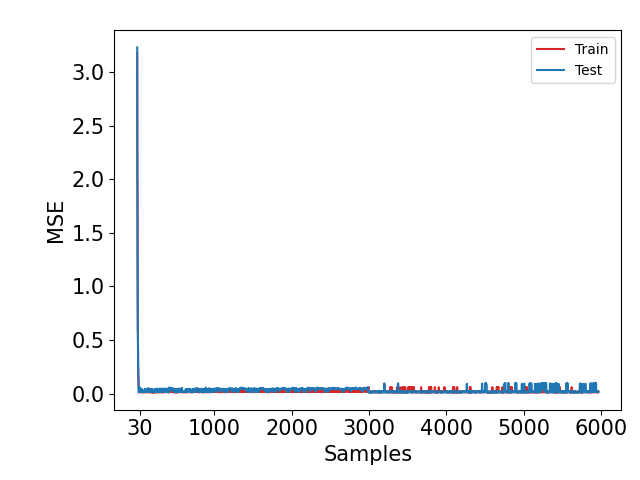}
 \caption{Madelon\label{subfig:RA_Madelon}}
 \end{subfigure}
 \begin{subfigure}{0.3\textwidth}
 \centering
 \includegraphics[scale = 0.33]{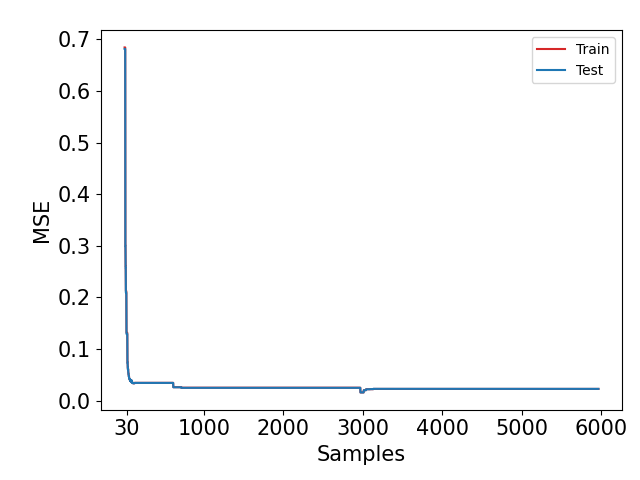}
 \caption{Coil 2000\label{subfig:RA_Coil}}
 \end{subfigure}
 \begin{subfigure}{0.3\textwidth}
 \centering
 \includegraphics[scale = 0.33]{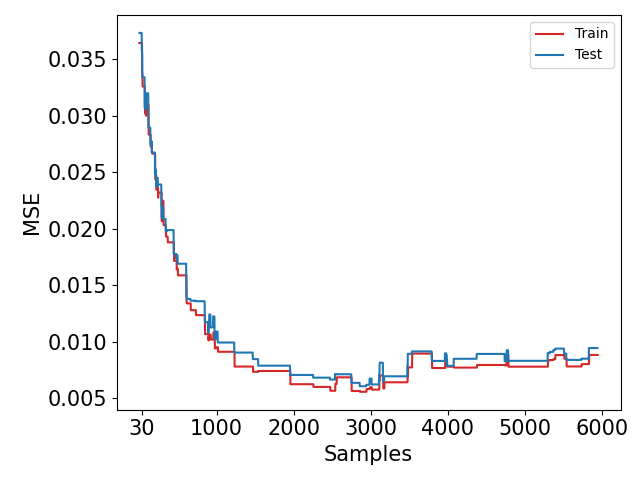}
 \caption{Swiss Roll}
 \end{subfigure}
 \caption{Reconstruction accuracy (MSE) over time (samples) for selected replica (temperature=1).}
 \label{fig:recon_samples}
\end{figure*}

\begin{figure*}[htbp!]
 \centering
 \begin{subfigure}{0.3\textwidth}
 \centering
 \includegraphics[scale = 0.33]{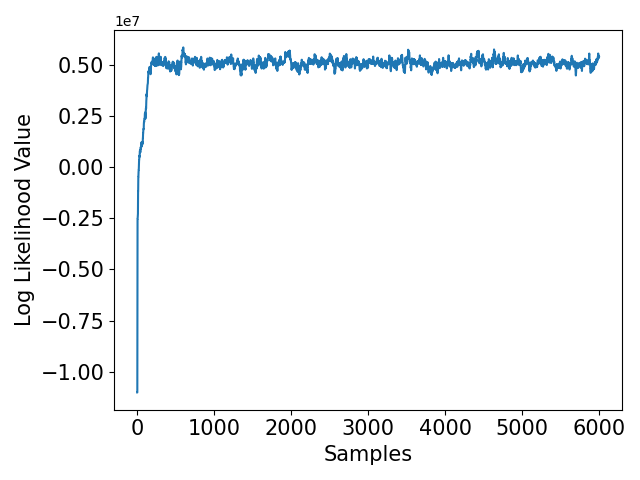}
 \caption{Madelon\label{subfig:l_Madelon}}
 \end{subfigure}
 \begin{subfigure}{0.3\textwidth}
 \centering
 \includegraphics[scale = 0.33]{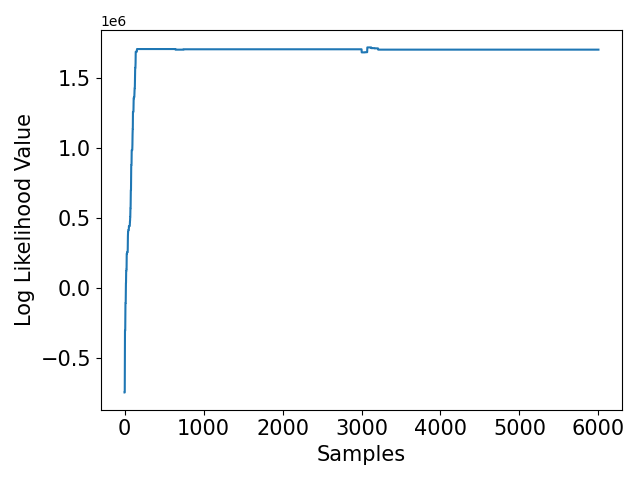}
 \caption{Coil 2000\label{subfig:l_Coil}}
 \end{subfigure}
 \begin{subfigure}{0.3\textwidth}
 \centering
 \includegraphics[scale = 0.33]{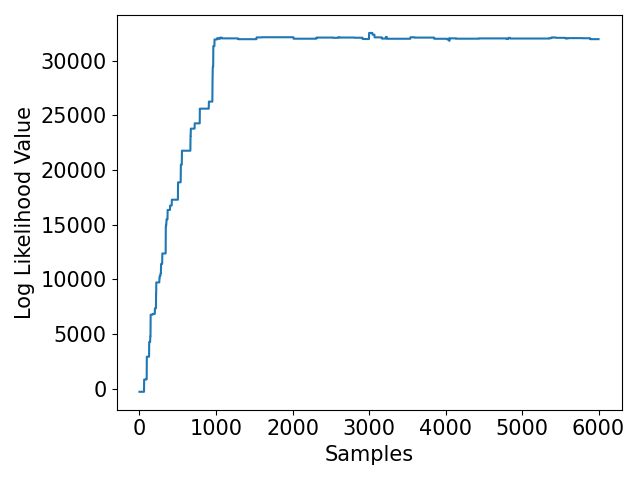}
 \caption{Swiss Roll\label{subfig:l_Swiss}}
 \end{subfigure}
 \caption{Log-Likelihood for replica with temperature=1}
 \label{fig:likelihood}
\end{figure*}

\begin{figure*}[htb]
 \centering
 \begin{subfigure}{0.3\textwidth}
 \centering
 \includegraphics[scale = 0.25]{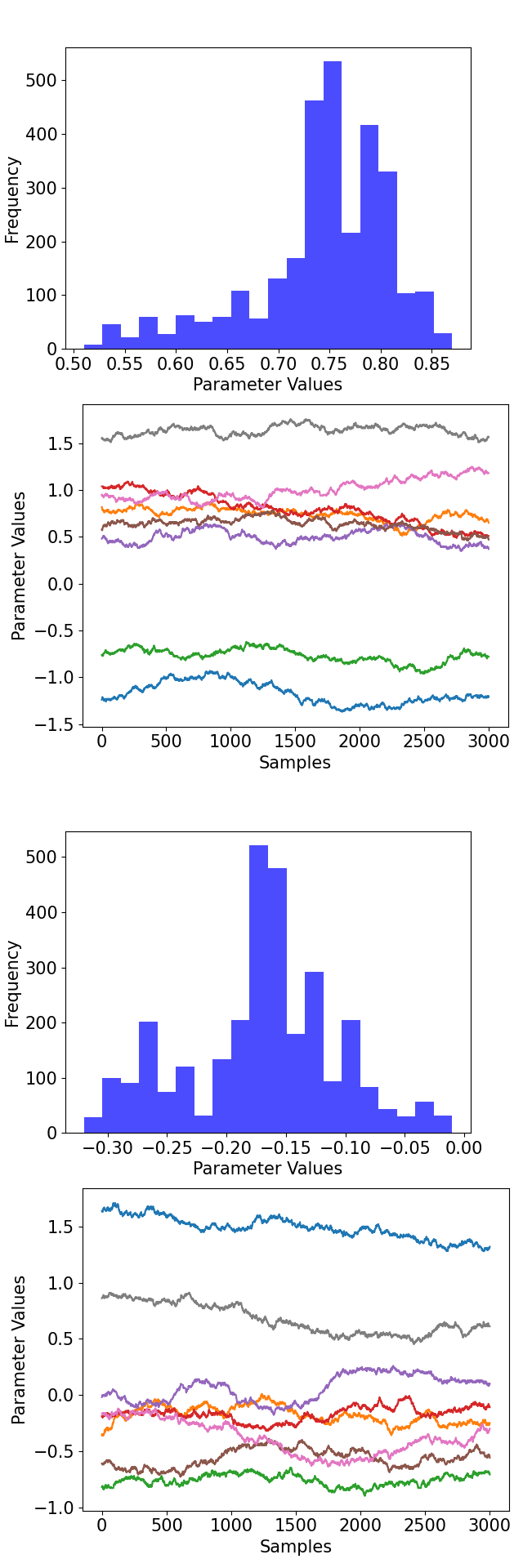}
 \caption{Madelon\label{subfig:posterior_Madelon}}
 \end{subfigure}
 \begin{subfigure}{0.3\textwidth}
 \centering
 \includegraphics[scale = 0.25]{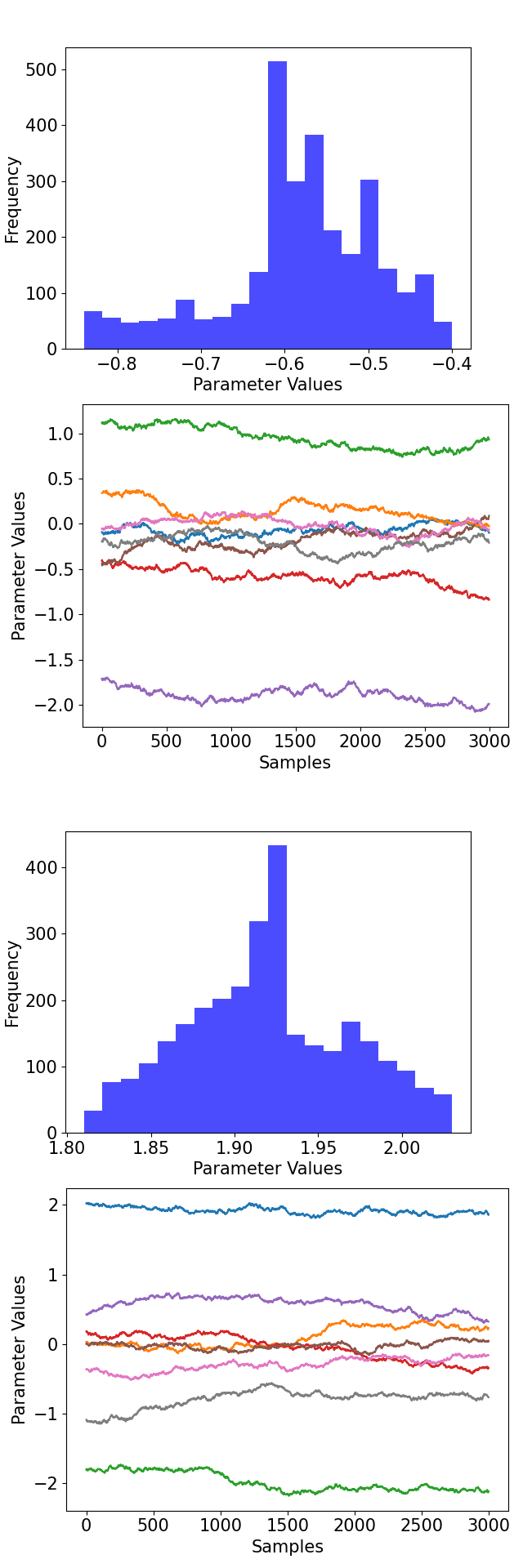}
 \caption{Coil 2000\label{subfig:posterior_Coil}}
 \end{subfigure}
 \begin{subfigure}{0.3\textwidth}
 \centering
 \includegraphics[scale = 0.25]{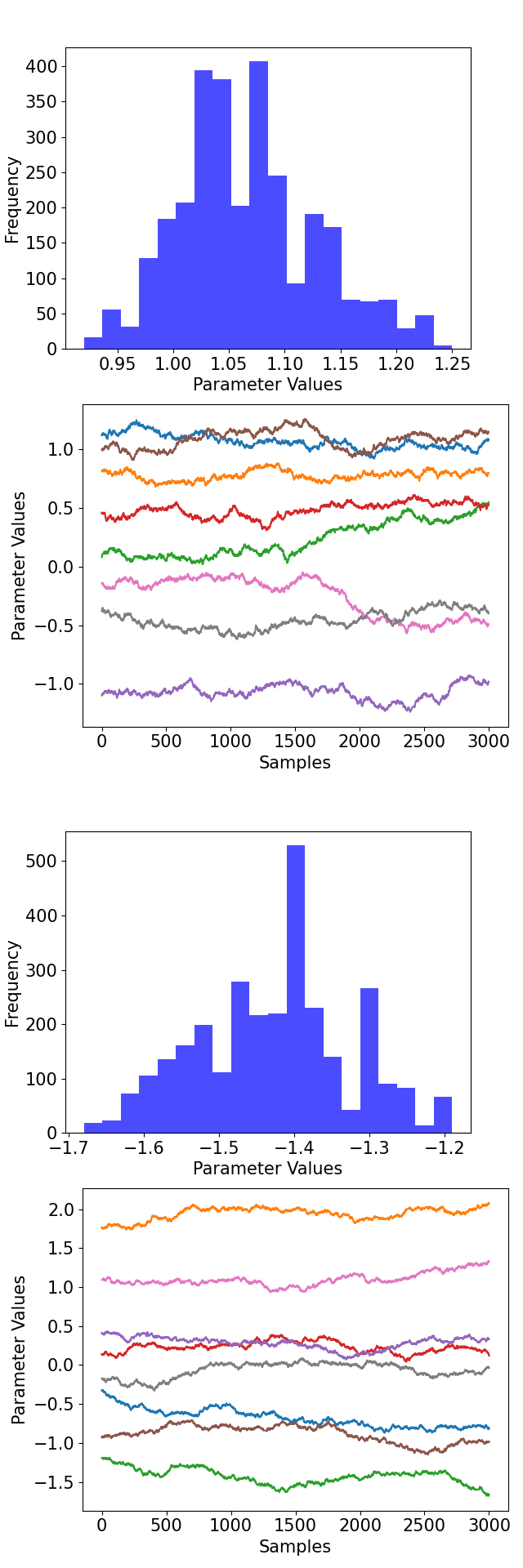}
 \caption{Swiss Roll\label{subfig:posterior_Swiss}}
 \end{subfigure}
 \caption{Posterior and trace plot for selected weights for the respective problems. }
 \label{fig:posterior}
\end{figure*}

 Figure~\ref{fig:recon_samples} shows trace plots over the samples with training and testing accuracy for one of the selected replicas of the Bayesian autoencoder (adapt-LG*), demonstrating convergence to the neighborhood of a posterior mode. We present a quantitative assessment of convergence in Table~\ref{tab:convergence} in the form of Gelman--Rubin \cite{vats2020revisiting} diagnostics. This diagnostic works by comparing the variance between replicates to the variance within them.  A ratio sufficiently close to 1 indicates convergence to the MCMC sampler's stationary distribution -- the posterior \cite{vats2020revisiting}. Figure~\ref{fig:likelihood} presents the log-likelihood of selected replicas (temperature level of 1.0) of the Bayesian autoencoder (adapt-LG*) for the respective problems. Figure~\ref{fig:posterior} shows the trace plots and posterior distribution for the selected weights of the Bayesian autoencoder (adapt-LG*) with the convergence diagnostic given in Table~\ref{tab:convergence}.

The data compression by autoencoders is typically bench-marked by applying machine learning methods on the reduced dataset.
We trained the Bayesian autoencoder and obtain a reduced dataset. Next, we train another machine learning model for the task of classification with reduced data obtained from the Bayesian autoencoder as done in previous studies \cite{ClassificationLit,classification_lit2, EnsembleClass, inproceedings,singh2020multiobjective, Autoencoder-MSELoss}. In this way, we can compare the performance by training the machine learning model on the reduced dataset with that of the original dataset. 

 Table~\ref{tab:compare_alg} presents a comparison of the results to selected methods from the literature that use the compressed datasets for classification tasks. We note that caution should be exercised when making direct comparisons, because the autoencoder architectures may vary somewhat between papers. To the extent that these comparisons are valid, our Bayesian autoencoder combined with SVM  outperforms the classification accuracy given in most methods from literature for the respective datasets. This shows the effectiveness of the Bayesian autoencoder as an alternative for dimensionality reduction and motivates the use of autoencoders as a possible data pre-processing step to increase classification accuracy.

 Table~\ref{tab:compare_MSE} compares the canonical autoencoder (using SGD and Adam optimizer) with the Bayesian autoencoder  (using LG and adapt-LG*). We show the best,  mean, and the standard deviation (std) of the reconstruction accuracy (MSE). As evident in Table~\ref{tab:compare_alg}, we find that the canonical autoencoder with SGD optimizer does not perform well for the respective datasets. We also find that the Bayesian autoencoder with adapt-LG* performs better than LG   for all three datasets. 

\begin{table*}[htbp!]
\centering
\small
 \caption{Gelman--Rubin MCMC convergence diagnostics for the weight chains in parallel tempering}
\label{tab:convergence}
\begin{tabular}{c c c c c c c c c c c}
\hline
Dataset & ID-0 & ID-50 & ID-100 & ID-150 & ID-2000 & ID-3000 & ID-4000 & ID-6000 & ID-9000 & ID-10,000\\
\hline
Madelon
& 1.15
& 1.14
& 1.15
& 1.21
& 1.16
& 1.13
& 1.18
& 1.25
& 1.13
& 1.25
\\
Coil 2000
& 1.14
& 1.16
& 1.18
& 1.23
& 1.15
& 1.17
& 1.11
& 1.24
& 1.12
& 1.13
\\
Swiss Roll
& 1.14
& 1.24
& 1.13
& 1.16
& -
& -
& -
& -
& -
& -
\\
\hline
\end{tabular}
\end{table*}

\begin{table*}
\caption{ Comparison of Bayesian autoencoder with established methods from literature. }
\label{tab:compare_alg}
\small
\centering
 \begin{tabular}{ c c c c }
 \hline
 Method & Algorithm & Madelon & Coil 2000 \\

 & & Best (Mean, Std) & Best (Mean, Std)\\
 \hline
 Autoencoder based classifier \cite{ClassificationLit} & kNN & 0.547 (-,-) & 0.929 (-,-) \\
 Ensemble Classification using Dimensionality Reduction \cite{EnsembleClass} & RSE & 0.5565 (- , 0.0263) & - \\
 Autoencoder inspired unsupervised feature selection \cite{inproceedings} & kNN & 0.71 (-,-) & - \\
 Multi-objective Evolutionary Approach \cite{singh2020multiobjective} & Naive Bayes & - & 0.93 ( -, 0.014) \\
 Multi-objective Evolutionary Approach \cite{singh2020multiobjective} & SVM & - & 0.948 ( -, 0.002)\\
 \hline
  Classifier (without compression) & SVM & 0.6 (0.56, 0.02 ) & 0.96 (0.95, 0.005)\\
 Bayesian-Autoencoder (LG) & kNN & 0.49 (0.45, 0.01 ) & 0.89 (0.81, 0.05) \\
 Bayesian-Autoencoder (adapt-LG*) & kNN & 0.554 (0.523, 0.018 ) & 0.958 (0.92, 0.015 ) \\
 Bayesian-Autoencoder (adapt-LG*) & SVM & 0.589 (0.554, 0.021 ) & 0.96 (0.94, 0.009 )
 \\
 \hline
\end{tabular}
\end{table*}

\begin{table*}
\caption{ Autoencoder reconstruction accuracy (MSE) for the three datasets. }
\label{tab:compare_MSE}
\small
\centering
 \begin{tabular}{c c c c }
 \hline
 Method & Swiss Roll & Madelon & Coil 2000 \\

 & Best (Mean, Std) & Best (Mean, Std) & Best (Mean, Std)\\
 \hline
Canonical Autoencoder (SGD) & 0.084 (0.087, 0.01) & 0.0389 (0.042,0.02) & 0.044 (0.0469, 0.03) \\
Canonical Autoencoder (Adam) &
0.012 (0.019, 0.005) & 0.0192 (0.0199, 0.007) & 0.0117 (0.014,0.002) \\
Bayesian Autoencoder (LG) & 0.020 (0.03,0.016) & 0.050 (0.060,0.003) & 0.033 (0.036,0.001) \\
Bayesian Autoencoder (adapt-LG*) &
0.010 (0.014,0.063) & 0.013 (0.023,0.006) & 0.016 (0.023,0.002) \\
\hline
\end{tabular}
\end{table*}

\section{Discussion} \label{sec:discussion}

Our primary goal was to incorporate Bayesian inference via MCMC sampling into autoencoders. The proposed Bayesian autoencoder provides an approach for uncertainty quantification in the reduced dataset. We can reconstruct an ensemble of reduced datasets from the posterior distribution, rather than a single one, given by single-point estimates. The results show that the Bayesian autoencoder framework can efficiently sample the posterior distribution of the autoencoder's weights and biases, which are compared very well with canonical methods in terms of accuracy (such as SGD and Adam). We also found that our approach is competitive with the other methods found in the literature.

It is unusual to use MCMC methods for deep learning, which often features models with tens of thousands of parameters; however, the field is slowly gaining momentum \cite{chandra2019langevin}. We have shown that with parallel computing and sophisticated gradient-based proposal distributions in MCMC, it is feasible for them to be used, nonetheless. We finally demonstrated how MCMC diagnostics apply to this problem (Table 6).

One limitation is that convergence analyses for MCMC samplers, such as the Gelman-Rubin diagnostics are generally designed for modest numbers of parameters. More work is needed to develop convergence diagnostic methods that are more suited for deep learning models that feature tens of thousands to millions of parameters. Furthermore, in large deep learning models, not all nodes and weights are equally valuable.  We note that a significant portion of these model parameters (weights) are not contributing, and hence they can be pruned during or post sampling (training). Optimization and evolutionary computation methods have been used for pruning weights of large deep learning models; hence, there is potential for Bayesian inference methods to be used in model pruning. The future research can focus on determining and differentiating essential weights from non-essential ones using specific metrics to determine which weights have a major effect on the performance of the model. Furthermore, a Bayesian framework can be developed to prune the non-essential weights that do not affect the performance of the model to build a robust and compact Bayesian autoencoder.

 We further highlight that we utilized uncorrelated multivariate Gaussian distributions as prior distribution for the autoencoder model parameters (weights and biases).  Although we consider previous research for selecting this prior distribution, we note that our previous research used different neural network architectures, i.e. simple neural networks \cite{Chandra2019NC}, and graph CNNs \cite{chandra2021bayesiangraph}. The recent developments on prior elicitation in Bayesian deep learning have spread the interest in more general choices \cite{fortuin2021priors}.  The choice of prior is a fundamental issue when the interest lies in performing Bayesian model selection \cite{llorente2020marginal}.  Further extensions of the proposed Bayesian autoencoder can consider current developments of priors in the field.

\section{Conclusions} \label{sec:conclusion}

We presented a framework  that employs tempered MCMC sampling for the Bayesian autoencoder which is an alternative  to the variational autoencoder. We addressed some of the known limitations of MCMC methods for Bayesian deep learning  with the power of parallel computing and enhanced proposal distributions. Our results indicate that the Bayesian framework is as good as related methods from the literature in terms of performance accuracy, with the additional feature of robust model uncertainty quantification for the generation of reduced datasets. The paper motivates further applications of the Bayesian framework for other deep learning models, such as the generative adversarial networks. Furthermore, there is also scope for domain-specific applications such as   remote sensing and geoscience applications where robust uncertainty quantification is vital.

\clearpage
\section*{Code and Data}

We provide the code as open-source software via our GitHub repository   \footnote{\url{https://github.com/sydney-machine-learning/BayesianAutoencoders}}.

\bibliographystyle{IEEEtran}
\bibliography{2017,2018,Chandra-Rohitash,Bays,sample,cnnbayes,autoenc}

\EOD

\end{document}